\documentclass[11pt]{article}

\usepackage{acl}
\usepackage{times}
\usepackage{latexsym}
\usepackage{graphicx}
\usepackage{amsmath}
\usepackage{amssymb}
\usepackage{booktabs}
\usepackage{multirow}
\usepackage{url}
\usepackage{listings}
\usepackage{xcolor}
\usepackage{subcaption}
\usepackage{enumitem}
\usepackage{array}

\title{\texttt{mllm-shap}: A Shapley Value Explainability Platform\\for Text-Audio Multimodal Large Language Models}

\author{
  Jakub Muszy\'{n}ski\textsuperscript{1}\footnotemark[1] \quad
  Pawe{\l} Pozorski\textsuperscript{1}\thanks{~~Equal contribution.} \quad
  Maria Ganzha\textsuperscript{1} \\
  \textsuperscript{1}Warsaw University of Technology, Warsaw, Poland \\
  \texttt{\{jakub.muszynski2.stud, pawel.pozorski.stud, maria.ganzha\}@pw.edu.pl} \\
  \small
  \href{https://orcid.org/0009-0000-2797-6044}{\includegraphics[height=8pt]{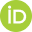} 0009-0000-2797-6044 (JM)} \quad
  \small \href{https://orcid.org/0009-0007-0728-1043}{\includegraphics[height=8pt]{images/orcid.png} 0009-0007-0728-1043 (PP)} \quad
  \small \href{https://orcid.org/0000-0001-7714-4844}{\includegraphics[height=8pt]{images/orcid.png} 0000-0001-7714-4844 (MG)}
}

\begin{document}
\maketitle

\begin{abstract}
We present \texttt{mllm-shap}, an open-source Python platform for researchers and ML practitioners that extends Shapley value (SV) explainability from text-only large language models to multimodal LLMs (MLLMs) that jointly process text and audio.
Building on the token-level SV framework introduced by TokenSHAP, \texttt{mllm-shap} addresses three challenges absent in the text-only setting: (1)~modality-aware coalition masking that handles the coexistence of text tokens and dense audio encoder frames within a single input, (2)~multi-turn conversation tracking with per-token role and modality metadata, and (3)~audio token grouping via phonetic alignment that reduces the coalition space by 10--50$\times$.
The platform ships as a \texttt{pip}-installable package\footnote{\url{https://pypi.org/project/mllm-shap/}} implementing five SV estimation strategies -- including a Complementary Contributions estimator with Neyman-optimal allocation that outperforms Monte Carlo baselines -- together with an interactive web GUI for real-time attribution visualization.
To our knowledge, \texttt{mllm-shap} is the first publicly available framework for complete, reproducible SV-based explainability of text-audio MLLMs. 
The package is MIT-licensed with full source code on GitHub\footnote{\url{https://github.com/Pawlo77/MLLM-Shap}}\textsuperscript{,}\footnote{\url{https://github.com/mvishiu11/shap-mllm-explainer}} and a demonstration video included as supplementary material.
\end{abstract}

\section{Introduction}
\label{sec:intro}

Shapley values \citep[SV;][]{RM-670-PR} have become a standard tool for model-agnostic input attribution in machine learning, popularized by SHAP \citep{lundberg2017unifiedapproachinterpretingmodel} for tabular and image models and extended to text-only LLMs by TokenSHAP \citep{goldshmidt2024tokenshapinterpretinglargelanguage}.
Unlike perturbation-based alternatives such as LIME \citep{ribeiro2016why} or gradient-based methods like Integrated Gradients \citep{sundararajan2017axiomatic}, SVs uniquely satisfy efficiency, symmetry, and additivity axioms while requiring only black-box access.
Yet as production systems increasingly deploy multimodal LLMs (MLLMs) that process both text and speech end-to-end -- in customer service, healthcare, and accessibility -- no existing tool supports SV-based explainability for these models \citep{zhao2023explainabilitylargelanguagemodels}.

Applying TokenSHAP's approach directly to text-audio MLLMs fails for three reasons.
First, \textbf{granularity mismatch}: a short utterance produces hundreds of dense audio encoder frames alongside a handful of text tokens, making the coalition space intractable ($2^{150+}$ vs.\ $2^{10}$).
Second, \textbf{modality-aware masking}: removing a text token means dropping it from the token sequence, but ``removing'' an audio segment requires zeroing encoder frames at specific time boundaries -- these are fundamentally different operations that must be handled within a single coalition.
Third, \textbf{conversational structure}: real MLLM interactions involve multi-turn dialogues with system prompts, user messages, and assistant responses across both modalities; attribution must track which turn and modality each token belongs to.

\texttt{mllm-shap} addresses all three gaps as an integrated platform with two main components:

\begin{enumerate}[itemsep=2pt]
    \item A \textbf{model-agnostic Python package} that extends SV computation to multimodal, multi-turn conversational inputs, with modality-aware masking, feature-unit metadata tracking, and five SV estimation strategies including a Neyman-optimal allocator (\S\ref{sec:package}).
    \item An \textbf{interactive web GUI} with cost estimation, session persistence, and real-time attribution visualization for both text and audio~(\S\ref{sec:gui}).
\end{enumerate}

We demonstrate the platform on LFM2-Audio-1.5B \citep{LIQUID_LFM2_AUDIO_2025}, showing concrete attribution outputs for text and multi-turn inputs (\S\ref{sec:walkthrough}), and report evaluation results across 593 SV analyses on a single consumer GPU (\S\ref{sec:eval}).

\section{System Architecture}
\label{sec:package}

\begin{figure}[ht]
    \centering
    \includegraphics[width=\columnwidth]{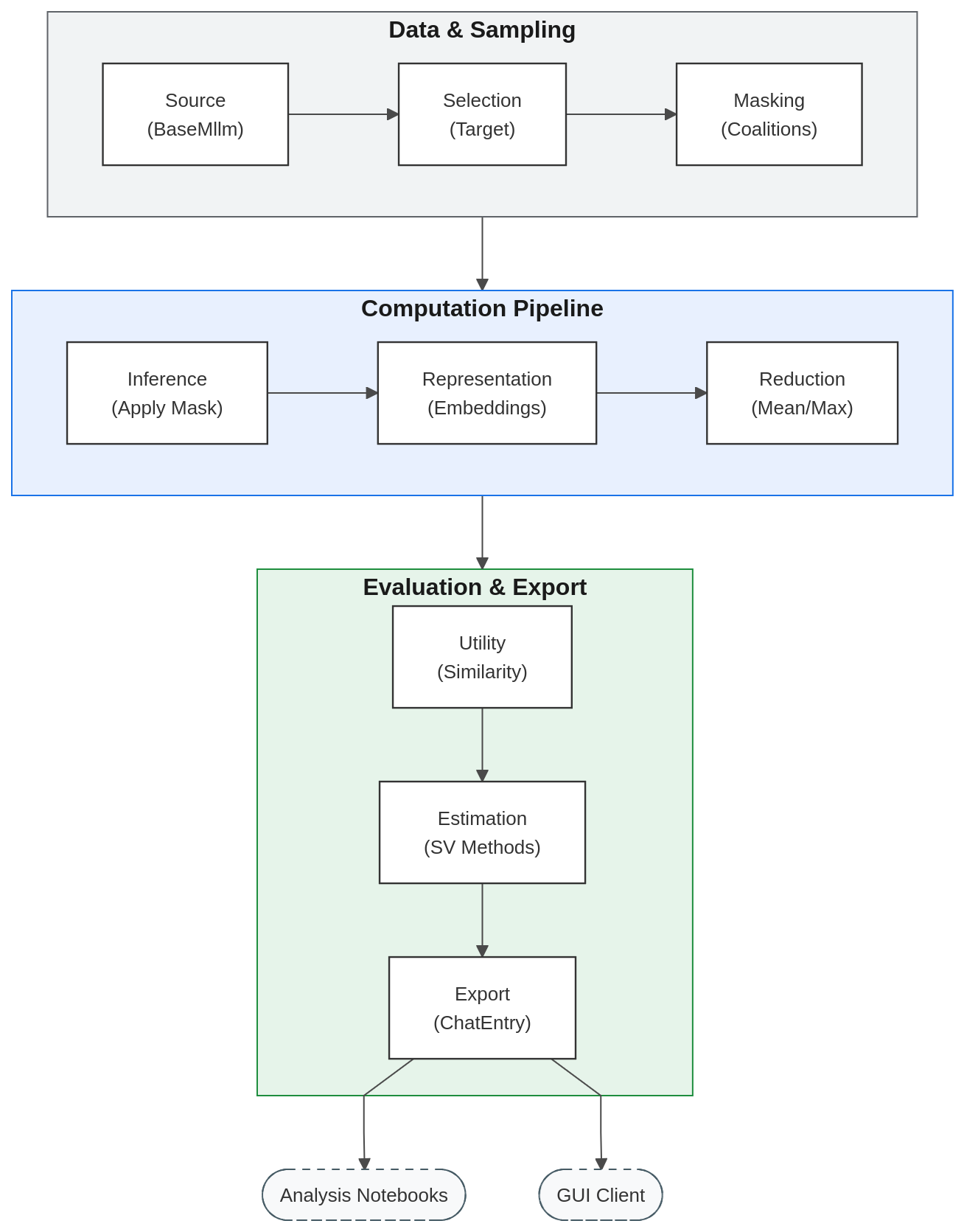}
    \caption{Package pipeline: chat construction with per-token metadata, modality-aware mask generation, connector-based inference, utility evaluation, and SV estimation. Results export to analysis notebooks or the GUI client.}
    \label{fig:arch}
\end{figure}

\texttt{mllm-shap} follows a modular, pipeline-oriented design (Figure~\ref{fig:arch}) that separates four concerns: (i)~model interaction via \emph{connectors}, (ii)~modality-aware coalition mask generation, (iii)~utility function evaluation, and (iv)~SV estimation.
This separation ensures that estimator comparisons differ only in the approximation method while all other components remain fixed.

\paragraph{Connectors and the Black-Box Interface.}
Model integration is encapsulated behind a \texttt{BaseMllmModel} interface that translates between package-level chat objects and model-specific I/O formats.
Only input--output access is required -- no gradients or attention maps -- making the system compatible with any MLLM exposing a generate API.
A reference connector for HuggingFace \texttt{transformers} is provided, with LFM2-Audio-1.5B as the default model.

\paragraph{Feature-Unit Tracking.}
A core design contribution is the \emph{feature unit}: each token in the pipeline carries structured metadata -- modality (\textsc{text} or \textsc{audio}), role (\textsc{user}, \textsc{system}, or \textsc{assistant}), turn index, and positional information.
This metadata flows through the entire pipeline, enabling modality-aware masking, role-based filtering (system tokens are excluded from attribution by default), and per-turn aggregation for multi-turn conversations.
Table~\ref{tab:multiturn} illustrates this: only user-content tokens (Role~0) receive SV values, while system/special tokens (Role~2) are correctly excluded, and turn indices enable separate analysis of each conversational exchange. Excluding tokens from coalitions within the scope of this framework means they are always fed to the model. That is, final input prompts consist of explainable tokens present in the given coalition (those with a non-NaN final SV) and all non-explainable tokens in their original order.

\paragraph{Modality-Aware Masking.}
A critical challenge in multimodal SV estimation is removing features in a way that remains in-distribution for the MLLM. The platform implements two distinct masking strategies based on feature type.
For \emph{text}, explainable tokens are physically removed from the input string; however, non-explainable role-markers (e.g., \texttt{<|im\_start|>}) and turn-delimiters are preserved in every coalition to maintain the structural integrity required by chat-tuned MLLMs, ensuring the model always perceives a valid conversational state.
For \emph{audio}, physical removal of segments would create temporal discontinuities and acoustic ``clicks'' that confuse the model's encoder; instead, \texttt{mllm-shap} masks by zeroing the amplitude of the raw waveform or latent frames within the time boundaries of the selected segment \citep{covert2021explaining}.
This dual-strategy design ensures that both modalities can participate in the same coalition game while each receives masking appropriate to its signal structure.\penalty-10000

\paragraph{Audio Token Grouping via SGPA.}

To resolve the granularity mismatch, the platform integrates Spectrogram-Guided Phonetic Alignment \citep{pozorski2026sgpa}, a four-stage pipeline that maps dense audio encoder frames to word-aligned segments.
SGPA combines CTC-based forced alignment \citep{graves2006connectionist} via Wav2Vec2-XLSR-53 \citep{baevski2020wav2vec} with spectrogram-guided boundary refinement using local energy and spectral flux cues, then aggregates character-level boundaries into word-level segments.
This reduces the effective coalition space by 10--50$\times$: in controlled experiments, SGPA compressed the player set from ${\sim}50$ native audio tokens to ${\sim}7$ word-aligned segments, yielding a 43$\times$ reduction in model evaluations and bringing per-sample wall-clock time from ${\sim}30$ minutes to roughly one minute on a consumer GPU.
Crucially, SGPA provides a model-agnostic explanation unit: by grounding attributions in phonetic segments rather than model-specific latent frames, \texttt{mllm-shap} enables direct, black-box explainability comparisons between models with different audio architectures.
We note that SGPA necessarily changes the cooperative game being solved -- attributions are defined with respect to externally aligned segments rather than model-native audio units -- and refer the reader to \citet{pozorski2026sgpa} for a full diagnostic characterization of this trade-off.

\begin{figure}[ht]
    \centering
    \includegraphics[width=\columnwidth]{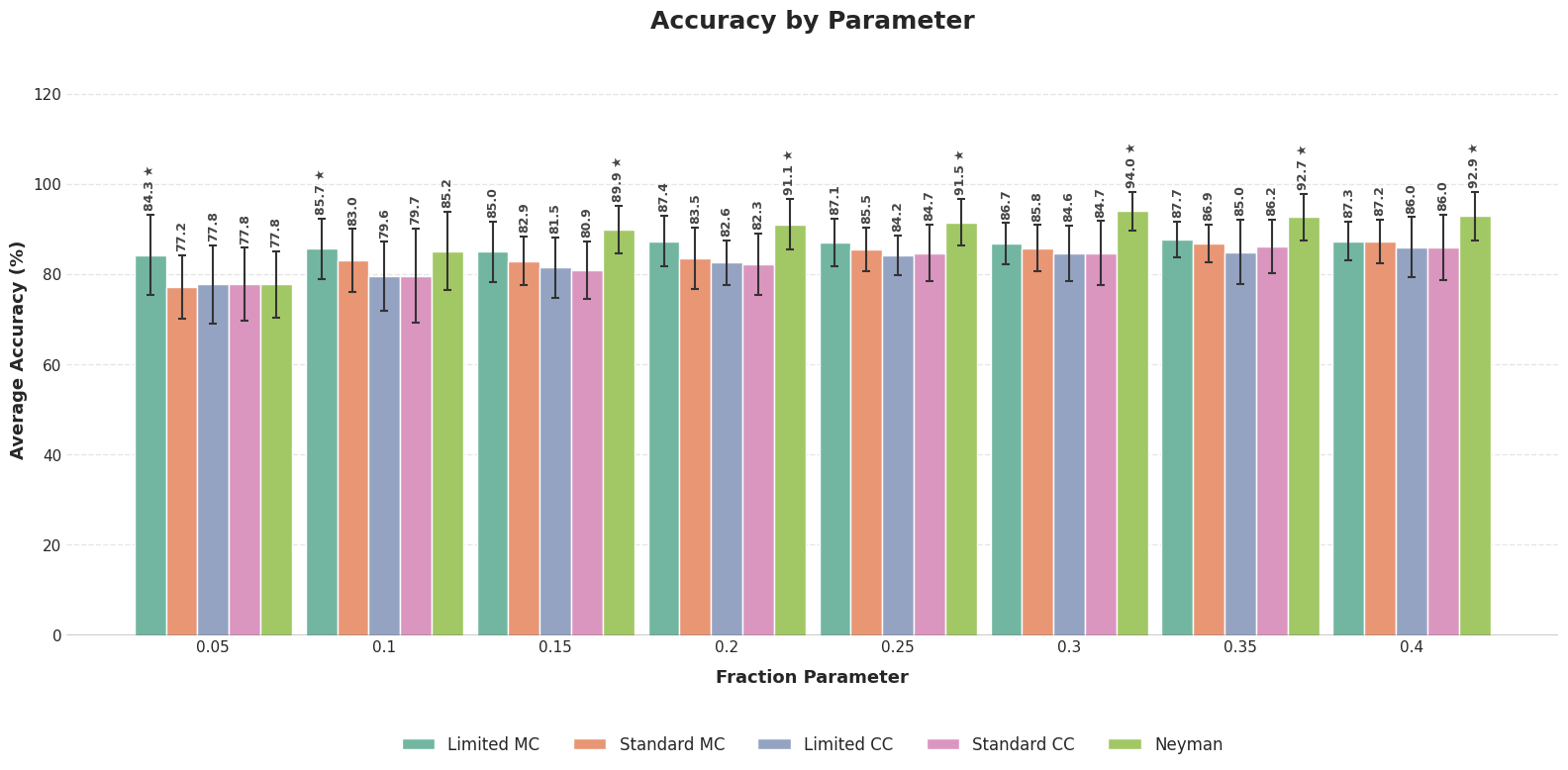}
    \caption{Approximation accuracy vs.\ exact SV (27 samples, 9--10 tokens). The Neyman-based estimator dominates at all sampling fractions tested.}
    \label{fig:benchmark}
\end{figure}

\paragraph{SV Estimators.}
The platform provides five estimation strategies sharing a common callable interface for seamless benchmarking: \textbf{Exact} ($2^n$ enumeration for tractable inputs), \textbf{Monte Carlo} with optional first-order omission constraints \citep{goldshmidt2024tokenshapinterpretinglargelanguage}, \textbf{Complementary Contributions} (CC) using stratified sampling \citep{shapleyapproximations}, \textbf{Neyman-CC} extending CC with Neyman-optimal variance-adaptive allocation \citep{nayman}, and \textbf{Hierarchical} recursive decomposition for long sequences.
Full algorithmic descriptions of each estimator are provided in Appendix~\ref{app:estimators}.
Benchmarking on 27 inputs where exact SV is tractable (9--10 tokens) confirms that Neyman-CC dominates, achieving $>90\%$ accuracy at sampling fractions as low as $0.15$ (Figure~\ref{fig:benchmark}).

\paragraph{Utility Functions.}
The platform supports three utility functions: Hidden State Similarity for white-box debugging, Cosine Similarity for semantic-level evaluation \citep{reimers}, and TF-IDF Weighted Cosine Similarity (the default) for cross-architecture benchmarking in a reduced feature space. Detailed descriptions are in Appendix~\ref{app:estimators}.

\section{Example Walkthrough}
\label{sec:walkthrough}

We illustrate the platform with two concrete examples on LFM2-Audio-1.5B, highlighting capabilities absent in existing SV tools.

\paragraph{Single-Turn Attribution.}
Listing~\ref{lst:example} shows a minimal API workflow.
For the text prompt \emph{``Who developed reCAPTCHA?''}, Table~\ref{tab:recaptcha} shows the resulting token-level SVs.
Note that the tokenizer splits ``reCAPTCHA'' into subword units (``Rec'', ``apt'', ``cha''); the platform handles this transparently, and the SVs reveal that ``developed'' ($\phi{=}0.110$) and the first subword of the target entity (``Rec'', $\phi{=}0.357$) carry high attribution -- consistent with the model attending to the verb--object pair that determines the answer.

\begin{table}[ht]
\centering
\small
\caption{Token-level SV for \emph{``Who developed reCAPTCHA?''} (text-to-text, LFM2-Audio-1.5B, Neyman estimator). System and punctuation tokens are automatically filtered.}
\begin{tabular}{clc}
\toprule
\textbf{ID} & \textbf{Token} & \textbf{SV ($\phi$)} \\
\midrule
1 & Who       & 0.080 \\
2 & developed & 0.110 \\
3 & Rec       & 0.357 \\
4 & apt       & 0.273 \\
5 & cha       & 0.174 \\
\bottomrule
\end{tabular}
\label{tab:recaptcha}
\end{table}

\begin{lstlisting}[caption={Minimal SV attribution with \texttt{mllm-shap}.},label=lst:example,float=t]
from mllm_shap.connectors.liquid import (
    LiquidAudio
)

from mllm_shap.shap.monte_carlo import LimitedMcShapExplainer
from mllm_shap.shap.compact import Explainer
from mllm_shap.connectors.enums import Role
import torch
from pprint import pprint

# Load model and create chat
model = LiquidAudio(device=torch.device("cuda"))
chat = model.get_new_chat()
chat.new_turn(Role.USER)
chat.add_text("Who developed Recaptcha?")
chat.end_turn()

# Run SV estimation
explainer = Explainer(
    model=model,
    shap_explainer=LimitedMcShapExplainer()
)
result = explainer(chat=chat)

pprint(result.full_chat.get_conversation())
\end{lstlisting}

\begin{figure}[ht]
    \centering
    \includegraphics[width=\columnwidth]{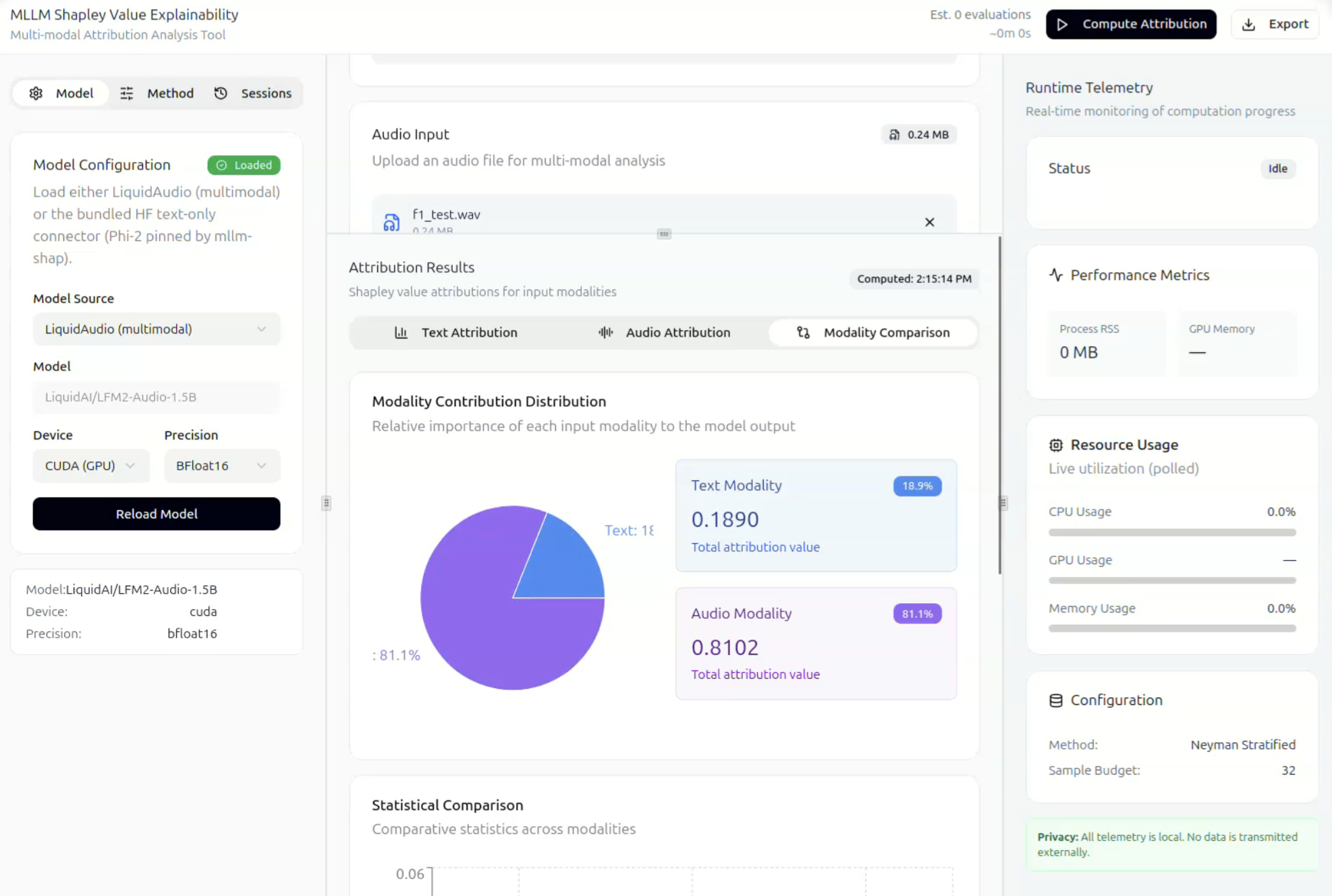}
    \caption{Web interface: chart of total SV attribution by modality based on simple text input of \emph{``Transcribe the audio.''} and $3$-second English audio recording.}
    \label{fig:multi-modal-example}
\end{figure}

\paragraph{Cross-Modal Attribution.}
To demonstrate multimodal reasoning, we analyze a transcription task where the input consists of a 3-second English audio recording and the text instruction: \emph{``Transcribe the audio.''}. As shown in Figure~\ref{fig:multi-modal-example}, the \textsc{audio} modality, represented by segments aligned via SGPA,  carries the primary attribution ($\phi{=}0.81$). In contrast, the text tokens for ``transcribe'' carry secondary attribution ($\phi{=}0.10$). The framework successfully decouples the source of information (the audio signal) from the instructional context (the text prompt).

\paragraph{Multi-Turn Conversation Attribution.}
Table~\ref{tab:multiturn} shows output from a two-turn dialogue: \emph{``Who are you?''} followed by \emph{``Can you repeat?''}.
This demonstrates three capabilities unique to \texttt{mllm-shap}:
(a)~\textbf{Turn tracking}: each token is tagged with its turn index, enabling per-turn and cross-turn attribution analysis.
(b)~\textbf{Role-based filtering}: special tokens (\texttt{<|im\_start|>}, \texttt{<|im\_end|>}) and role markers are assigned Role~2 (system) and automatically excluded from SV computation, while user-content tokens (Role~0) receive attributions.
(c)~\textbf{Structured output}: the full token--SV--role--turn table is exportable as a DataFrame for downstream statistical analysis.

In this example, ``Who'' ($\phi{=}0.231$) and ``you'' ($\phi{=}0.230$) dominate Turn~1, while ``are'' receives zero attribution -- the model's response to an identity question is driven by the interrogative and the pronoun, not the copula. In Turn~3, ``repeat'' ($\phi{=}0.217$) carries the highest SV, consistent with it being the only novel semantic content (the model must recognize a repetition request).

\begin{table}[ht]
\centering
\small
\caption{Multi-turn attribution for \emph{``Who are you? $|$ Can you repeat?''}. Role~0 = user content (attributed), Role~2 = system/special (excluded) -- denotes tokens excluded from SV computation by design.}
\begin{tabular}{clccr}
\toprule
\textbf{ID} & \textbf{Token} & \textbf{SV ($\phi$)} & \textbf{Role} & \textbf{Turn} \\
\midrule
0  & \texttt{<|im\_start|>} & ---    & 2 & 1 \\
1  & user                   & ---    & 2 & 1 \\
3  & Who                    & 0.2313 & 0 & 1 \\
4  & are                    & 0.0000 & 0 & 1 \\
5  & you                    & 0.2300 & 0 & 1 \\
7  & \texttt{<|im\_end|>}   & ---    & 2 & 1 \\
\midrule
9  & \texttt{<|im\_start|>} & ---    & 2 & 3 \\
10 & user                   & ---    & 2 & 3 \\
12 & Can                    & 0.1546 & 0 & 3 \\
13 & you                    & 0.1676 & 0 & 3 \\
14 & repeat                 & 0.2166 & 0 & 3 \\
16 & \texttt{<|im\_end|>}   & ---    & 2 & 3 \\
\bottomrule
\end{tabular}
\label{tab:multiturn}
\end{table}

\section{Interactive Web Interface}
\label{sec:gui}

\begin{figure}[ht]
    \centering
    \includegraphics[width=\columnwidth]{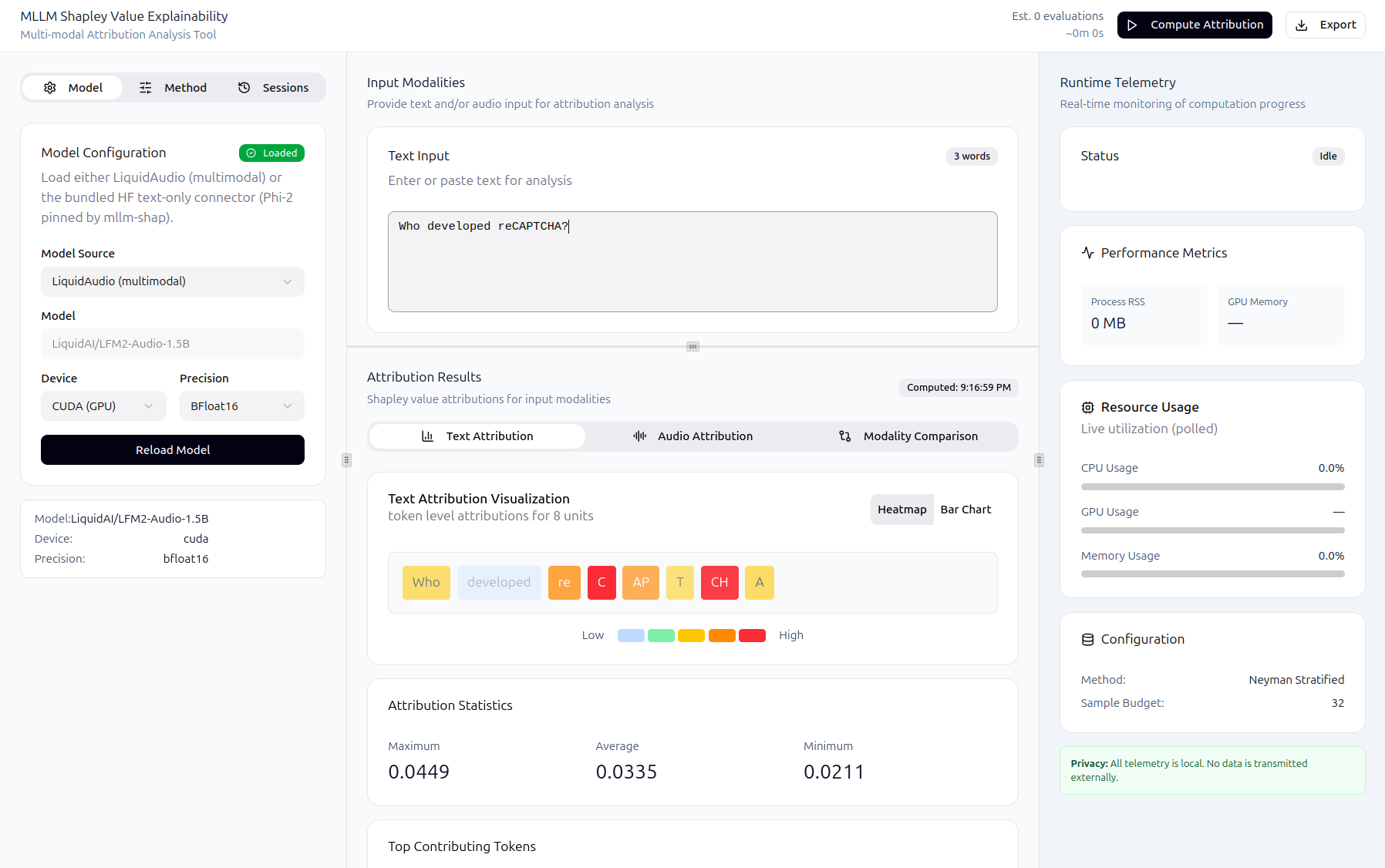}
    \caption{Web interface: model and method configuration (left), multimodal chat with text and audio input (center), attribution heatmaps and time-aligned audio plots (right).}
    \label{fig:gui}
\end{figure}

The GUI (Figure~\ref{fig:gui}) provides a complete explainability workflow without writing code.
It follows a client--server architecture: a React \citep{facebook2013react} frontend communicates with a FastAPI\footnote{\url{https://fastapi.tiangolo.com/}} backend via REST API, with PostgreSQL for session persistence and Nginx for routing.

\paragraph{Asynchronous Execution.}
Since SV estimation requires hundreds of model forward passes, the backend implements an asynchronous task queue using FastAPI's background workers.
When a user requests an explanation, the request is immediately acknowledged with a unique task ID and the estimation process is dispatched to a GPU-bound worker, preventing HTTP timeouts and allowing the frontend to poll for real-time progress updates visualized via a progress bar.

\paragraph{Session Persistence.}
The PostgreSQL database stores not only chat history but also the Feature-Unit metadata for every token and audio segment.
This enables \emph{session resumption} -- a researcher can close the browser during a long-running Neyman-CC estimation and return later to find results fully populated -- and \emph{post-hoc analysis}, where computed attributions can be re-visualized using different GUI modes (e.g., switching from a modality-level pie chart to a token-level heatmap) without re-running the computation.

\begin{figure}[ht]
    \centering
    \includegraphics[width=\columnwidth]{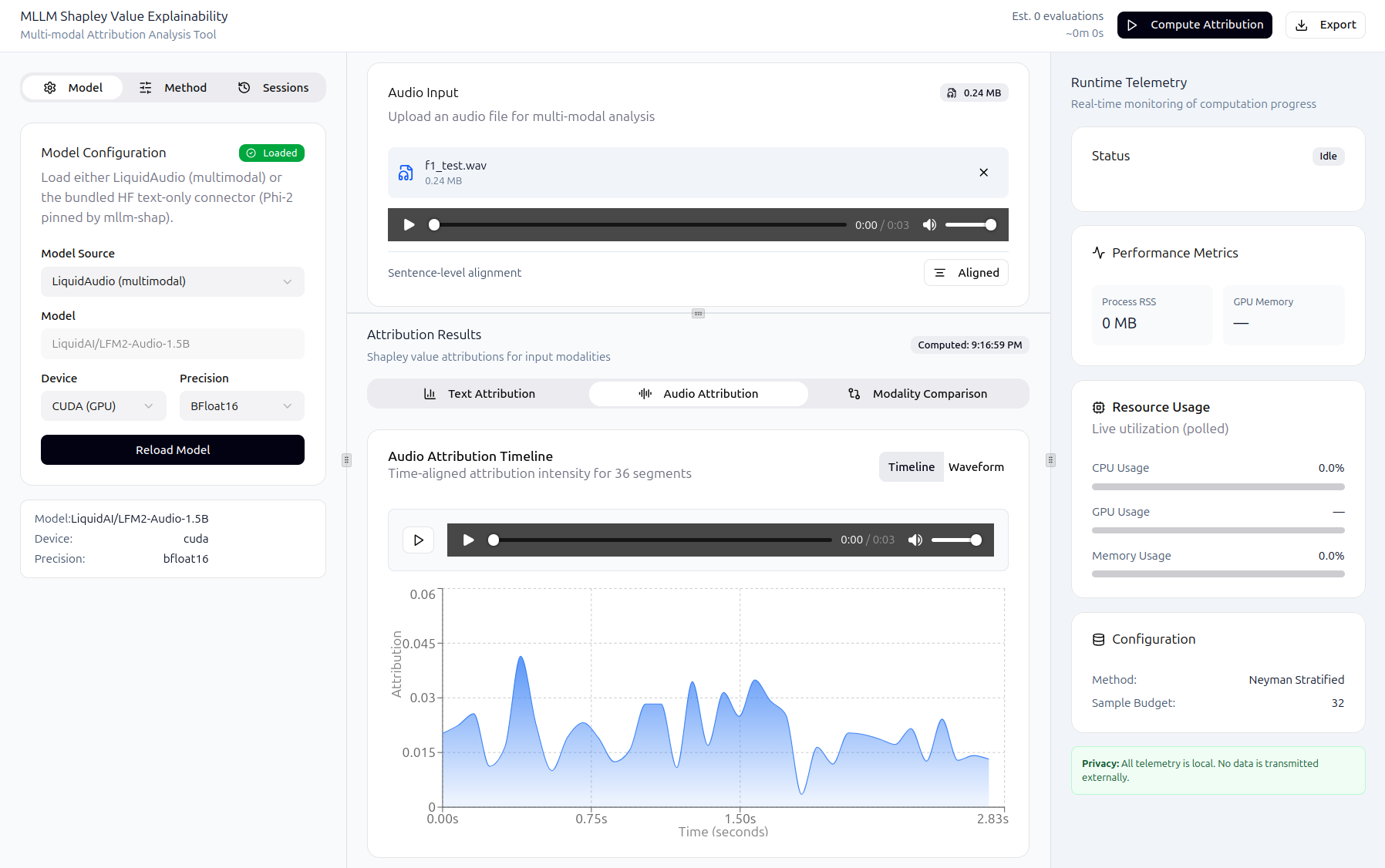}
    \caption{Audio attribution view: time-aligned SV intensity plot for a 3-second audio input, with model configuration (left), attribution results with audio/text/modality comparison tabs (center), and runtime telemetry with method configuration (right).}
    \label{fig:gui_audio}
\end{figure}

\paragraph{Workflow and Visualization.}
The interface guides users through the following steps: configure model and SV method $\to$ interact via text or audio $\to$ receive an upfront cost estimate (number of model calls and estimated runtime) $\to$ compute attributions $\to$ visualize results as token heatmaps or time-aligned audio plots $\to$ export for analysis.
For audio inputs, the GUI provides a dedicated \emph{Audio Attribution Timeline} (Figure~\ref{fig:gui_audio}) that displays time-aligned attribution intensity overlaid on the waveform, allowing researchers to identify which temporal segments of the audio signal most influenced the model's output.
The right panel provides real-time telemetry (CPU/GPU/memory utilization), the active SV configuration, and a privacy notice confirming that all computation remains local.
The GUI is containerized via Docker Compose (Appendix~\ref{app:gui_arch}) for reproducible single-command deployment.

\section{Evaluation}
\label{sec:eval}

We evaluate \texttt{mllm-shap} along three axes: estimator accuracy, platform-scale feasibility, and coverage of system capabilities.

\paragraph{Estimator Accuracy.}
On 27~samples where exact SV is tractable (9--10 tokens), the Neyman estimator with first-order omission achieves $>$90\% mean accuracy at 30\% sampling fraction ($\sim$150 coalitions), outperforming standard MC and CC at all tested budgets (Figure~\ref{fig:benchmark}).
Across all estimators, the modular pipeline ensures that comparisons isolate the approximation method: connector behavior, masking policies, and utility evaluation remain fixed, so accuracy differences reflect only the estimation strategy.

\paragraph{System Feasibility.}
We validated \texttt{mllm-shap} by performing 593 SV analyses on a single NVIDIA RTX 4080 GPU (16\,GB VRAM), covering 100 VoiceBench \citep{chen2024voicebench} samples across five modality configurations (text-to-text, speech-to-text with male/female voices, and speech-to-speech variants) and 93 multilingual samples from Infinity-Instruct \citep{li2025infinityinstruct} in three languages (English, Spanish, French).
Table~\ref{tab:feasibility} summarizes key runtime statistics.
Single-sentence audio-to-text tasks complete in $\sim$30\,s, while multi-sentence interleaved prompts, which expand the coalition space by 10--20$\times$, require~$\sim$400\,s.
The Neyman allocator consistently reached its variance-adaptive second phase in the majority of runs, confirming that even the default budget allocation ($3n^2$ model evaluations) provides sufficient initialization for stable estimates.
Speech-to-speech modes incur $\sim$2$\times$ the runtime of speech-to-text due to audio decoding overhead, establishing a practical ceiling for interactive use.

\begin{table}[ht]
\centering
\small
\caption{System feasibility summary across 593 SV analyses on a single consumer GPU. Runtime is mean wall-clock time per sample using the Neyman estimator.}
\begin{tabular}{lrr}
\toprule
\textbf{Configuration} & \textbf{Mean calls} & \textbf{Mean time (s)} \\
\midrule
Single-sentence S2T    & 60   & 30  \\
Single-sentence S2S    & 59   & 67  \\
Single-sentence T2T    & 155  & 91  \\
Multi-sentence (interleaved) & 1{,}005--1{,}174  & 399--451 \\
Multilingual S2T       & 235  & 105 \\
Multilingual T2T       & 606  & 205 \\
\bottomrule
\end{tabular}
\label{tab:feasibility}
\end{table}

\paragraph{Capability Coverage and Reproducibility.}
Across the 593 analyses, all core platform capabilities were exercised at scale: cross-modal attribution (300 analyses), SGPA audio segmentation (200), multi-turn tracking (100), and multilingual support across English, Spanish, and French (93) with no language-specific pipeline modifications required (Appendix~\ref{app:capabilities}).
The full results are released as a \textbf{multimodal attribution dataset} on HuggingFace\footnote{\url{https://huggingface.co/datasets/Pawlo77/mllm-shap}}, providing a reproducible baseline for future multimodal XAI research.
All experiments were executed through a reproducible pipeline with YAML-based configuration management, per-sample checkpointing with resume support, and deterministic inference defaults (temperature~=~0, top-$k$~=~1, see Appendix~\ref{app:infra}).

\section{Related Systems}
\label{sec:related}

Table~\ref{tab:comparison} compares \texttt{mllm-shap} with existing SV-based tools. While the original SHAP \citep{lundberg2017unifiedapproachinterpretingmodel} is the industry standard, it lacks native support for the variable-length, auto-regressive nature of LLMs. TokenSHAP \citep{goldshmidt2024tokenshapinterpretinglargelanguage} addressed text-only LLMs but cannot handle the ``interleaved'' nature of multimodal conversations or the high-dimensionality of raw audio encoder frames. Prior reviews of SHAP-based methods in NLP \citep{mosca2022shap} have similarly focused exclusively on text modalities, while attention-based interpretability approaches \citep{jain2019attention} remain debated in their faithfulness and do not naturally extend to cross-modal settings where text and audio operate over different representational spaces. \texttt{mllm-shap} fills this gap by introducing a \textbf{Conversation-aware Feature-Unit} framework, allowing for the first time a unified, principled SV analysis of how text instructions and audio signals cooperatively drive MLLM behavior.

\begin{table}[ht]
\centering
\small
\caption{Comparison with existing SV-based explainability tools.}
\begin{tabular}{lccc}
\toprule
\textbf{Feature} & \textbf{SHAP} & \textbf{TokenSHAP} & \textbf{Ours} \\
\midrule
Text LLMs           & \checkmark & \checkmark & \checkmark \\
Audio modality       & $\times$   & $\times$   & \checkmark \\
Variable-length seq. & $\times$   & \checkmark & \checkmark \\
Multi-turn chat      & $\times$   & $\times$   & \checkmark \\
Audio segmentation   & $\times$   & $\times$   & \checkmark \\
Neyman allocation    & $\times$   & $\times$   & \checkmark \\
Interactive GUI      & $\times$   & $\times$   & \checkmark \\
Reproducibility infra & $\times$  & $\times$   & \checkmark \\
Open-source          & \checkmark & \checkmark & \checkmark \\
\bottomrule
\end{tabular}
\label{tab:comparison}
\end{table}

\section{Conclusion and Availability}
\label{sec:conclusion}

We presented \texttt{mllm-shap}, the first open-source platform for Shapley value-based explainability of text-audio multimodal language models.
By extending the token-level SV framework of TokenSHAP with modality-aware masking, multi-turn conversation tracking, efficient audio token grouping via SGPA, and an interactive GUI, the system makes previously intractable multimodal attribution analysis accessible on consumer hardware.
Future work includes extending connector support to additional MLLM architectures (e.g., vision-language models), implementing Shapley interaction indices for relational attribution, and conducting large-scale human evaluation studies.

The platform is distributed under the MIT license.
The \texttt{pip}-installable package,\footnote{\textbf{PyPI:} \url{https://pypi.org/project/mllm-shap/}} source repositories,\textsuperscript{2,3} API documentation,\footnote{\url{https://pawlo77.github.io/MLLM-Shap/}} and the multimodal attribution dataset\textsuperscript{4} are all publicly available.
The interactive GUI requires a CUDA-capable GPU for model inference and cannot be hosted as a public live demo; a screencast video demonstrating the full workflow is included as supplementary material.

\section*{Ethical Considerations}
The system is designed to improve transparency and accountability of AI systems.
No personal or sensitive data is collected by the tool; all experiments used publicly available datasets and open-source models.
We note that SV-based attributions, while principled, should not be treated as ground-truth causal explanations of model behavior -- they measure marginal contribution under a specific utility function and coalition structure.
Additionally, SGPA segmentation introduces an external player partition that does not necessarily align with model-internal representations.

\section*{Limitations}
All experiments use a single MLLM (LFM2-Audio-1.5B); although the connector interface is model-agnostic, generalization to other architectures (e.g., Whisper-based or codec-based speech models) remains unvalidated.
Despite SGPA's 43$\times$ coalition-space reduction, complex multi-sentence multimodal inputs still require ${\sim}400$\,s per sample on consumer hardware, limiting real-time interactivity for long inputs.
Only text and audio modalities are currently supported; vision or video would require new masking and alignment strategies.
Finally, the evaluation measures system feasibility and estimator accuracy against exact SVs but does not include human studies of attribution interpretability or downstream utility.

\setlength{\bibsep}{0pt plus 0.3ex}
\bibliography{references}

\appendix

\section{SV Estimator and Utility Function Details}
\label{app:estimators}

\paragraph{Estimator Descriptions.}
The five SV estimation strategies provided by \texttt{mllm-shap} are:

\begin{itemize}[itemsep=1pt]
    \item \textbf{Exact:} Full $2^n$ enumeration of all coalitions, restricted to tractable inputs (typically $n \le 15$). Serves as the ground-truth reference for estimator benchmarking.
    \item \textbf{Monte Carlo (MC):} Random coalition sampling with optional first-order omission constraints, following \citet{goldshmidt2024tokenshapinterpretinglargelanguage}. Omission constraints ensure that each player is excluded from at least one sampled coalition, reducing the variance of individual SV estimates.
    \item \textbf{Complementary Contributions (CC):} A stratified sampling approach where each coalition evaluation contributes to the marginal utility of all included players and their complements simultaneously \citep{shapleyapproximations}, approximately doubling the information extracted per model call.
    \item \textbf{Neyman-CC:} An extension of CC that utilizes Neyman-optimal allocation~\citep{nayman} to distribute the sampling budget across strata according to their variance, significantly reducing estimation error for a fixed budget. The estimator operates in two phases: an initial uniform exploration phase ($3n^2$ evaluations by default) followed by a variance-adaptive allocation phase.
    \item \textbf{Hierarchical:} A recursive decomposition strategy designed for long sequences that exceed the computational limits of flat estimators. Tokens are grouped into higher-level features, SVs are computed at the group level, and then recursively refined within groups of interest.
\end{itemize}

\paragraph{Utility Functions.}
The platform supports three utility functions tailored to different research goals:
\textbf{(U1)~Hidden State Similarity} directly compares model-internal activations between full and masked inputs, enabling white-box debugging of how coalitions affect latent representations.
\textbf{(U2)~Cosine Similarity} computes sentence-level semantic similarity using external embeddings \citep{reimers}, providing a model-external measure of output equivalence.
\textbf{(U3)~TF-IDF Weighted Cosine Similarity} operates over decoded text, weighting terms by their TF-IDF scores so that attribution focuses on the information-dense tokens that define the model's factual output. U3 is the system default, as its term-weighting makes it optimal for cross-architecture benchmarking where internal representations are inaccessible.

\section{GUI Deployment Architecture}
\label{app:gui_arch}

\begin{figure}[ht]
    \centering
    \includegraphics[width=0.85\columnwidth]{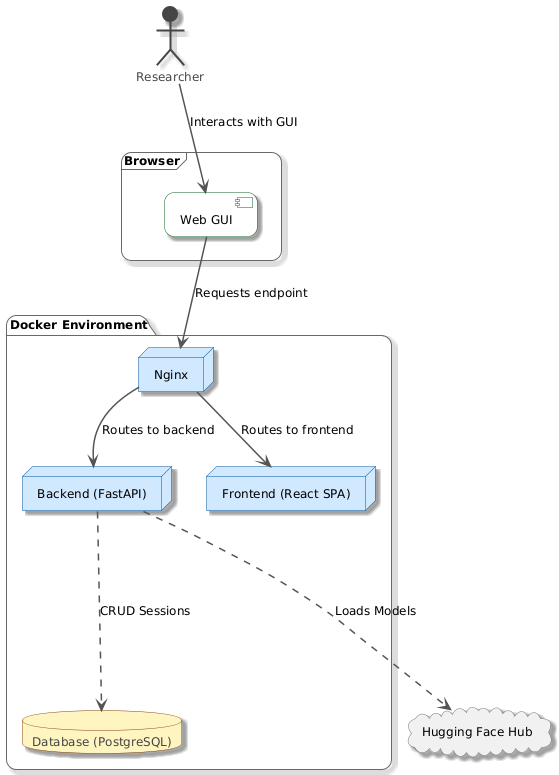}
    \caption{GUI deployment architecture: three-container Docker Compose setup with Nginx routing, FastAPI backend loading models from HuggingFace Hub, and PostgreSQL for session CRUD.}
    \label{fig:gui_arch}
\end{figure}

The system architecture (Figure~\ref{fig:gui_arch}) is orchestrated via Docker Compose to manage the lifecycle of three distinct services: an Nginx reverse proxy for routing, a FastAPI backend that loads models from HuggingFace Hub and performs GPU-bound SV computation, and a PostgreSQL database for session CRUD operations. The GUI is containerized for reproducible single-command deployment on any system with a compatible NVIDIA Container Toolkit installation.

\section{Experimental Pipeline and Reproducibility}
\label{app:infra}

\begin{figure}[ht]
    \centering
    \includegraphics[width=0.85\columnwidth]{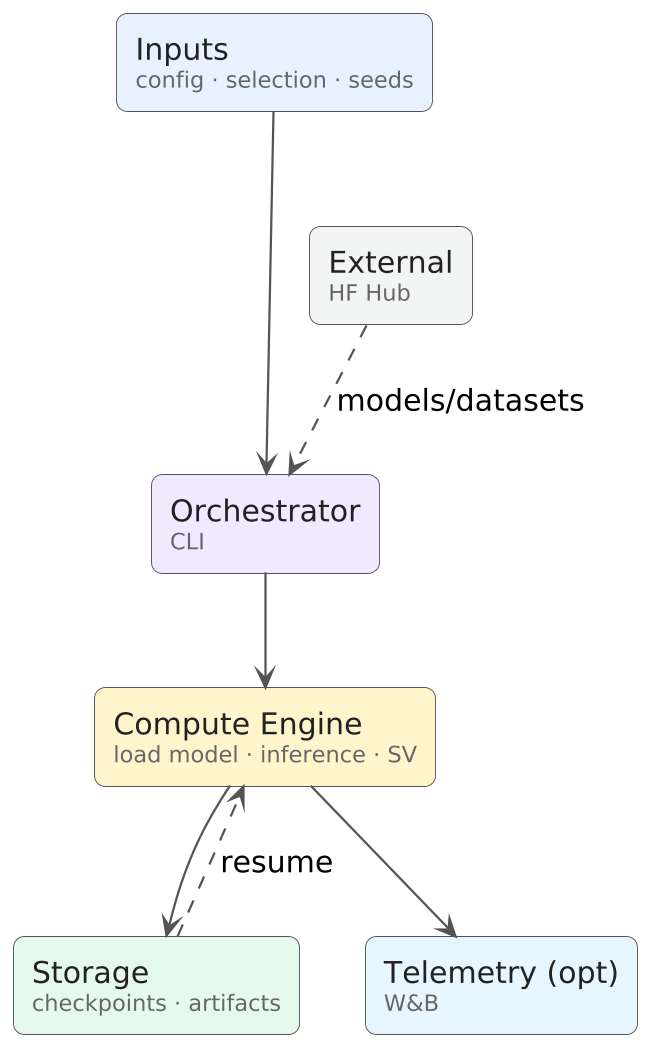}
    \caption{Experimental infrastructure: YAML-configured orchestrator dispatches jobs to the GPU compute engine with per-sample checkpointing.}
    \label{fig:infra}
\end{figure}

To support the 593 analyses reported in Section~\ref{sec:eval}, we developed a robust experimental orchestrator (Figure~\ref{fig:infra}) organized around four key components:
\textbf{Configuration Management} uses YAML-based specifications to pin model versions, estimator hyperparameters, and dataset revisions.
The \textbf{Deterministic Compute Engine} enforces deterministic inference settings (temperature~=~0, top-$k$~=~1).
A \textbf{Checkpointing Layer} persists results incrementally, enabling resume from the last completed sample if a run is interrupted.
The \textbf{Artifact Hierarchy} produces a structured output directory per run containing a \texttt{spec.json}, checkpoint states, and individual JSON results for every sample with full metadata and attribution vectors.
All evaluation experiments were conducted on a single consumer-grade workstation equipped with an NVIDIA RTX 4080 (16\,GB VRAM) and an Intel Core i9-13900K CPU.

\section{Capability Coverage}
\label{app:capabilities}

Table~\ref{tab:capabilities} provides a detailed breakdown of the platform capabilities exercised across the 593-analysis evaluation suite. Each capability corresponds to a distinct engineering challenge solved by the platform: multi-turn tracking was validated on the full VoiceBench multi-sentence set (4 modes $\times$ interleaved prompts); cross-modal attribution was exercised across all speech configurations; and multilingual support was confirmed across three typologically diverse languages.

\begin{table}[ht]
\centering
\small
\caption{Platform capabilities exercised across the evaluation suite. Each row corresponds to a system feature validated at scale.}
\begin{tabular}{lrl}
\toprule
\textbf{Capability} & \textbf{Analyses} & \textbf{Source} \\
\midrule
Text-only attribution     & 200 & VB + II \\
Cross-modal (text+audio)  & 300 & VoiceBench \\
SGPA audio segmentation   & 200 & VoiceBench \\
Multi-turn tracking       & 100 & VB multi-sent. \\
Multilingual support      & 93  & Infinity-Inst. \\
\bottomrule
\end{tabular}
\label{tab:capabilities}
\end{table}

\section{Attribution Data Schema}
\label{app:schema}

The platform exports results in a structured JSON format (Table~\ref{tab:json_schema}) that encapsulates the complete state of the Shapley game. For large-scale sweeps, the platform provides a CLI orchestrator that consumes these schemas to generate aggregate reports, with support for parallelization across multiple GPUs.

\begin{table}[ht]
\centering
\small
\caption{Structure of the exported attribution JSON.}
\begin{tabular}{llp{4.5cm}}
\toprule
\textbf{Field} & \textbf{Type} & \textbf{Description} \\
\midrule
\texttt{session\_id} & UUID & Unique identifier for the analysis session. \\
\texttt{metadata}   & Dict & Model name, estimator type, and budget. \\
\texttt{features}   & List & Array of Feature-Units (ID, modality, role). \\
\texttt{values}     & List & Raw $\phi$ values corresponding to each unit ID. \\
\texttt{turn\_map}  & Dict & Mapping of turn indices to feature IDs. \\
\texttt{utility}    & Float & Final utility score of the grand coalition. \\
\texttt{telemetry}  & Dict & GPU memory usage and wall-clock time. \\
\bottomrule
\end{tabular}
\label{tab:json_schema}
\end{table}

\end{document}